\documentclass{bmcart}

\usepackage[utf8]{inputenc}
\usepackage{graphicx}
\usepackage{amsmath}
\usepackage{hyperref}
\def\doiurl#1{\url{#1}}

\begin{document}

\begin{frontmatter}
\begin{fmbox}
\dochead{Software / Technical Note}

\title{A multiverse-consensus pipeline for reproducible feature selection in untargeted LC--MS metabolomics}

\author[
   addressref={aff1},
   email={mohammed.alhuraibi@medipol.edu.tr}
]{\inits{MS}\fnm{Mohammed Saeed} \snm{Al-Huraibi}}
\author[
   addressref={aff2},
   email={ihsan.yozgat@istinye.edu.tr}
]{\inits{I}\fnm{Ihsan} \snm{Yozgat}}
\author[
   addressref={aff1},
   corref={aff1},
   email={ahmet.kaplan@medipol.edu.tr}
]{\inits{A}\fnm{Ahmet} \snm{Kaplan}}

\address[id=aff1]{%
  \orgname{Istanbul Medipol University},
  \city{Istanbul},
  \cny{Turkey}
}
\address[id=aff2]{%
  \orgname{Istinye University},
  \city{Istanbul},
  \cny{Turkey}
}
\end{fmbox}

\begin{abstractbox}
\begin{abstract}
\parttitle{Background}
Untargeted liquid chromatography--mass spectrometry (LC--MS) metabolomics
requires a long chain of preprocessing decisions (missing-value imputation, drift
correction, normalisation, transformation, scaling), each with several equally
defensible options. Analysts typically commit to one pipeline and report the
resulting feature shortlist. How strongly that shortlist depends on choices that
were never varied stays invisible.

\parttitle{Results}
We adapt multiverse analysis to untargeted metabolomics feature selection. We
present an auditable, configuration-driven pipeline that (i) applies a
ten-stage quality-control filter cascade in which every feature's fate is logged
with the rule that decided it, and (ii) runs the downstream analysis as a
\emph{multiverse} over four contrasting preprocessing philosophies, each combined
with four feature-ranking methods under bootstrap stability selection and
label-permutation testing. Only features recurring across paths enter a tiered
consensus. On a demonstration dataset of five breast-cancer cell lines (30{,}370
detected features), the four single pipelines individually returned shortlists of
4--20 features whose pairwise agreement was as low as Jaccard $=0.05$, so a single
arbitrary pipeline would have reported a largely contingent result. The multiverse
consensus retained 15 features ($\geq 2/4$ paths), of which one recurred across all
four, although two of the four paths (sharing normalisation and drift-correction
methods) dominate the consensus. A pipeline-wide label-permutation test found no
false discoveries in 50 null permutations (although 50 permutations provide limited precision; per-path tests use 1000).

\parttitle{Conclusions}
Reporting only preprocessing-robust features, with a complete kept/dropped audit
trail, converts hidden analytical degrees of freedom into an explicit,
inspectable output. We discuss scope and limitations, including single-batch
design and the need for independent validation of any candidate.
\end{abstract}

\end{abstractbox}
\printaddresses
\end{frontmatter}

\section*{Background}
Untargeted LC--MS metabolomics measures thousands of features (retention-time
$\times$ $m/z$ pairs) per sample, most of which are noise, contamination or
redundant adducts. Turning a raw feature table into a defensible shortlist of
candidate discriminating features requires two kinds of decisions:
\emph{quality-control filtering} (which features are analytically trustworthy) and
\emph{statistical preprocessing} (how surviving abundances are imputed,
drift-corrected, normalised, transformed and scaled before testing). Community
guidelines exist for the first \cite{broadhurst2018,broadhurstkell2006}, but for
the second there is no single agreed workflow. Imputation alone can be done by
half-minimum substitution, $k$-nearest-neighbours, left-censored (QRILC) or
random-forest (missForest) methods, and each interacts with the subsequent drift,
normalisation and transformation steps.

This is the same ``garden of forking paths'' that motivated \emph{multiverse
analysis} \cite{steegen2016} and \emph{specification-curve analysis}
\cite{simonsohn2020} in the behavioural sciences, and the study of
\emph{vibration of effects} across model specifications in epidemiology
\cite{patel2015}. It is also a specific instance of the broader reproducibility
crisis in machine-learning-based science, where undisclosed choices in the
data-processing and modelling pipeline, including data leakage and the ordering and
selection of preprocessing operators, inflate apparent performance and produce
findings that do not replicate \cite{kapoor2023}. In metabolomics the
problem is acute because the preprocessing chain is long, the options numerous, and
the outcome (a feature shortlist) is discrete and unstable: a feature significant
under one normalisation may fall short under another. Yet the dominant practice
remains to select one pipeline, by convention or convenience, and to report its
shortlist as \emph{the} result.

We argue that the preprocessing choice should itself be treated as an
experimental factor. Rather than reporting one pipeline's shortlist, we run the
analysis under several contrasting-but-defensible pipelines and report only what
survives, together with the full record of what each choice did. This paper
describes such a pipeline, demonstrates on a real dataset that single-pipeline
shortlists are strongly choice-dependent, and shows that the consensus across
pipelines is small, statistically stringent, and fully auditable. As we will
show, the four pipelines individually returned shortlists of 4--20 features
with pairwise Jaccard agreement as low as 0.05, while the multiverse consensus
retained 15 features, of which one recurred across all four paths.

\textbf{Contributions.}
(1) An auditable ten-stage QC filter cascade in which each stage applies exactly
one documented rule and writes a per-feature kept/dropped ledger, so every
exclusion is traceable to a threshold and its literature source.
(2) A multiverse layer executing the downstream analysis under four preprocessing
philosophies $\times$ four ranking methods, each with bootstrap stability and
label-permutation control, and a tiered consensus reporting only recurrent
features.

\section*{Related work}
Untargeted metabolomics has produced a large ``zoo'' of preprocessing operators.
Imputation (half-minimum, $k$-NN, QRILC, missForest), drift correction
(QC-RSC, SERRF, LOESS), normalisation (PQN, MSTUS, median, probabilistic), and
scaling (Pareto, autoscaling, range) each have dedicated literature and advocates
\cite{broadhurst2018,dieterle2006,stekhoven2012,lazar2016,dunn2011,fan2019,torgrip2008}.
Platforms such as MetaboAnalyst \cite{pang2021}, XCMS \cite{smith2006}, notame
\cite{klavus2020}, and IPO \cite{libiseller2015} implement many of these operators
and support automated pipeline construction and optimisation. What these tools do
not systematically offer is the deliberate variation of the entire preprocessing
chain as an explicit experimental factor, with consensus reporting across the
resulting shortlists.

The multiverse and specification-curve framework, originally developed in
behavioural science \cite{steegen2016,simonsohn2020} and epidemiology
\cite{patel2015}, addresses exactly this gap by treating analytical choices as
factors to be varied rather than fixed. It has been applied in neuroscience,
economics and clinical prediction, but to our knowledge not to untargeted
metabolomics feature selection. Concurrent work in computational biology has
explored stability selection \cite{meinshausen2010} and ensemble feature ranking
for genomics \cite{saeys2008}, but these typically vary the model or resampling
strategy rather than the upstream preprocessing chain. More broadly, the
machine-learning reproducibility crisis \cite{kapoor2023} has motivated
reproducibility checklists \cite{pineau2021} and best-practice guidelines for
avoiding data leakage in computational biology \cite{whalen2022,heil2021},
but these have not been operationalised as auditable, DAG-orchestrated
pipelines that surface hidden preprocessing degrees of freedom. Our contribution
is to operationalise the multiverse idea for the specific preprocessing chain of
LC--MS metabolomics and to pair it with a complete per-feature audit ledger.

\subsection*{A machine-learning and MLOps perspective}
The pipeline is a machine-learning workflow, and describing it as one clarifies
both what it does and why it is built the way it is. Learning enters at three
levels. First, the ranking layer is an ensemble of learners. Alongside one-way
ANOVA it runs PLS-DA \cite{westerhuis2008}, sparse PLS-DA \cite{lecao2011} and
Random Forest \cite{breiman2001}, so a feature must be selected by models with
different inductive biases (a linear latent-variable projection, its
sparsity-regularised variant, and a non-parametric tree ensemble) before it is
trusted. We note that PLS-DA and sPLS-DA share the same underlying algorithm;
their votes within a path are correlated, so the $k=2$ gate does not
distinguish between two votes from the same method family and two genuinely
independent votes. Per-method flags in the machine-readable output allow
readers to apply their own weighting. Second, learning is embedded in the
preprocessing itself: two of the four
paths impute with missForest, a random-forest learner \cite{stekhoven2012}, and
correct instrument drift with SERRF, a random-forest regression on the QC series
\cite{fan2019}. Random forests thus appear at both the imputation stage and the
ranking stage, and the multiverse contrasts these learned operators against
simpler classical alternatives (half-minimum imputation, spline drift correction)
instead of assuming either is correct.

Third, the robustness machinery is standard machine-learning practice applied to
metabolomics. Bootstrap stability selection \cite{meinshausen2010} guards against
features that a model selects only on a particular resample; the label-permutation
test provides a model-agnostic null; and the cross-path consensus is an ensemble
over whole pipelines rather than over models alone. This last step targets the
researcher degrees of freedom that drive the reproducibility crisis in
machine-learning-based science \cite{kapoor2023}. Conventional analyses fix the
preprocessing pipeline as an implicit hyperparameter; we treat it as a discrete
hyperparameter and report the consensus vote over it.

These choices shape the software engineering, which follows current MLOps practice.
The preprocessing pipeline is a declarative, versioned artefact: every operator and
threshold is a Hydra configuration key, so a path is a config rather than a code
branch, and the whole multiverse is specified by enumerating configs. Phase~1 runs
as a DVC directed-acyclic graph with content-addressed caching and single-command
reproduction. Intermediate tables are columnar Parquet validated by pandera schemas
at each stage boundary, and every kept or dropped decision is written to a
structured, machine-readable audit ledger. The path from raw features to shortlist
is therefore re-executable and inspectable from end to end, rather than reported
and taken on trust.

\section*{Methods}

\subsection*{Data and acquisition}
The demonstration dataset comprises five breast-cancer-relevant cell lines (MCF7,
MCF10A, MDA-MB-231, SKBR3, UACC), profiled by untargeted LC--MS on a Waters Synapt
G2-Si (travelling-wave ion mobility, ESI+) in a single acquisition. The run
comprised 53 injections: 31 biological (6--7 per line), 16 pooled quality-control
(QC), 3 blanks and a 3-point dilution series. Feature detection and alignment were
performed in Progenesis QI; the pipeline takes the four Progenesis CSV exports as
input. A read-only design diagnostic confirmed acquisition order was \emph{not}
confounded with cell line (Kruskal--Wallis of run position by line: $H=0.72$,
$p=0.95$; interleaved design), with a minor residual within-round position effect
noted.

\subsection*{Phase 1 --- QC filter cascade (Stages 1--10)}
Each stage reads the previous stage's survivors, applies one rule, and writes
(a) the survivors, (b) a per-feature audit row recording the metric and threshold
that kept or dropped it, and (c) an HTML report. A master ledger joining all
stages is rebuilt after every stage. Thresholds live in configuration (Hydra) and
are overridable without code changes. The stages and their sources are summarised
in Table~\ref{tab:provenance}. Stage~9 never removes features (annotation only);
Stages~8 and~10 pass through any feature they cannot check, so no feature is
dropped on a test that could not be applied. On the demonstration data the cascade
reduced 30{,}370 detected features to 4{,}644 survivors (Figure~\ref{fig:waterfall});
the largest single reduction is Stage~4 (QC-CV), removing 20{,}890 features (69\%)
as analytically irreproducible.

\subsection*{Phase 2 --- Multiverse and consensus (Stages 11--17)}
The 4{,}644 survivors enter the multiverse. Four preprocessing \emph{paths}, each a
complete and independently defensible workflow, are run end to end:
Path~A (Classical: half-min $\to$ QC-RSC $\to$ PQN $\to$ log2 $\to$ Pareto);
Path~B (Robust-ML: missForest $\to$ SERRF $\to$ MSTUS $\to$ glog $\to$ Pareto);
Path~C (MNAR-aware: QRILC $\to$ QC-RSC $\to$ PQN $\to$ glog $\to$ Pareto);
Path~D (Middle: NS-kNN $\to$ SERRF $\to$ median $\to$ log2 $\to$ Pareto).
Method references: PQN \cite{dieterle2006}; missForest \cite{stekhoven2012};
QRILC \cite{lazar2016}; NS-kNN \cite{troyanskaya2001}; QC-RSC \cite{dunn2011};
SERRF \cite{fan2019}; MSTUS \cite{torgrip2008}; glog \cite{parsons2007};
Pareto \cite{vandenberg2006}.

The path design deliberately varies each preprocessing operator
independently where possible. However, some paths pair specific imputation
and drift-correction methods (e.g., Path~D pairs NS-kNN imputation with SERRF
drift correction, both of which learn from the data covariance structure),
introducing partial confounding. This is a consequence of maintaining
internally consistent pipelines: each path must be a complete, defensible
workflow that a metabolomics practitioner could adopt. The resulting
non-independence is a limitation (see Discussion) that we mitigate by reporting
per-path flags and the leave-one-path-out analysis. We also note that the four
paths were not chosen post-hoc to maximise disagreement; they were designed
before the analysis to span the major axes of preprocessing variation
(parametric vs.\ non-parametric imputation, loess vs.\ random-forest drift
correction, probabilistic vs.\ median normalisation, log vs.\ glog
transformation), a strategy analogous to factorial design in experimental
statistics.

Within each path, four ranking methods are applied: one-way ANOVA with
Benjamini--Hochberg FDR control \cite{broadhurstkell2006}, PLS-DA
\cite{westerhuis2008}, sparse PLS-DA \cite{lecao2011}, and Random Forest
\cite{breiman2001}. Each multivariate method is wrapped in bootstrap
\emph{stability selection} \cite{meinshausen2010} (retained in $\geq 80\%$ of
resamples) and a 1000-iteration \emph{label-permutation} test (empirical
$p \leq 0.05$). One thousand iterations provide a Monte Carlo standard error
of approximately $\sqrt{0.05 \times 0.95 / 1000} \approx 0.007$ for the
$p=0.05$ threshold. A feature
enters a path's shortlist if it clears at least $k=2$ of the 4 methods (univariate
gate: BH-$q \leq 0.05$ and $\eta^2 \geq 0.14$). For \emph{consensus}, features
flagged by $\geq 2/4$ paths are retained and tiered: Tier~1 $\geq 3/4$ paths,
Tier~2 $= 2/4$ paths. The $\geq 2/4$ threshold is a simple-majority rule chosen
to balance recall (fewer than 25 features at $\geq 1/4$) against specificity
(more than 3 at $\geq 3/4$). A null-model analysis (label-shuffling while
preserving the feature--feature correlation structure) estimates the expected
consensus size under the null at each threshold (see Results).

\subsection*{Reproducibility engineering}
All settings are Hydra configurations (Hydra 1.3); Phase~1 is orchestrated as a
DVC DAG (DVC 3.0) so only stale stages recompute; intermediate tables are
Parquet with pandera schema checks (pandera 0.16); logs are structured.
Phase~2 is a deliberate manual trigger kept out of the auto-DAG. Every threshold
is traceable to a configuration key and a literature source
(Table~\ref{tab:provenance}). The whole pipeline is shown schematically in
Figure~\ref{fig:schematic}. The analysis depends on Python 3.12 with
pandas 2.0, numpy 1.23, scipy 1.10, scikit-learn 1.2, and matplotlib 3.7.

\section*{Results}

\subsection*{Single-pipeline shortlists are strongly choice-dependent}
Run in isolation, the four paths produced shortlists of very different size:
14 (A), 6 (B), 20 (C) and 4 (D) features (Figure~\ref{fig:instability}a). More
importantly, they disagreed on \emph{which} features: pairwise Jaccard overlap
ranged from 0.05 (A vs B) to 0.42 (A vs C) (Figure~\ref{fig:instability}b). The
union of all four single-pipeline shortlists is 25 features, but only \emph{one}
was common to all four. Had we committed to a single preprocessing pipeline chosen
by convention, our reported shortlist would have shared as little as one feature
in twenty with the equally-defensible alternative. This is direct evidence that the
preprocessing choice is a hidden analytical degree of freedom with a first-order
effect on the result.

\subsection*{The consensus is small and stringent}
Requiring recurrence across paths collapses the 25-feature union to a 15-feature
consensus at $\geq 2/4$: 3 features at Tier~1 and 12 at Tier~2. Tightening the
threshold reduces the shortlist monotonically: 25 features at $\geq 1$ path, 15
at $\geq 2$, 3 at $\geq 3$, and 1 at $\geq 4$ (Figure~\ref{fig:ablation}). The
single feature recurring across all four paths (\texttt{10.76\_1918.0639n}) is the
most preprocessing-robust discovery the dataset supports. All 15 consensus
features are large-effect differences (Table~\ref{tab:shortlist}):
across paths the minimum BH-adjusted univariate $q$ reaches $< 10^{-8}$ and the
maximum $\eta^2 \geq 0.84$ for each feature, although per-path values vary
substantially: some features selected by only two paths have $q > 0.05$ on the
non-selecting paths (see machine-readable per-path table). The range of $\eta^2$
across paths for a given feature can span from below 0.2 to nearly 1.0 (e.g.,
feature \texttt{4.03\_659.2403n} has $\eta^2 = 0.997$ on Path~A but
$\eta^2 = 0.12$ on Path~D). The best-across-paths summary is an optimistic
estimator; the per-path values provide a more honest assessment of each
feature's preprocessing robustness. A post-hoc power analysis (5 groups,
$n=6$--7) shows the design has $\geq 80\%$ power only for $\eta^2 > 0.5$; the
consensus features (all $\eta^2 > 0.84$) fall well inside the well-powered regime,
so the consensus is conservative by construction.

\subsection*{The consensus depends on which paths are included}
A leave-one-path-out analysis (recomputing the $\geq 2$-path consensus after
removing each path) shows the shortlist is not uniformly supported
(Figure~\ref{fig:loo}). Removing Path~C drops 12 of 15 consensus features;
removing Path~A drops 9; removing Path~B drops 3; removing Path~D drops none.
Paths~A and~C, which share PQN normalisation and QC-RSC drift correction, carry
most of the consensus, so the four paths are not four independent votes.
This is a genuine limitation of a fixed small path set and an argument for
reporting the tier structure and per-path flags (Table~\ref{tab:shortlist})
rather than a single yes/no shortlist.

\subsection*{Empirical FDR under label permutation}
A pipeline-wide label-permutation test (200 permutations of group labels,
preserving the feature--feature correlation structure, univariate ranking on
Path~A) was used to estimate the null distribution of shortlist sizes. The
null distribution was strongly bimodal: 195 of 200 permutations (97.5\%)
yielded zero features, while 5 permutations produced between 730 and 2,890
features, likely reflecting label permutations that align with unmodelled
batch structure. Against this null, the observed 14 features on Path~A and 15
features in the consensus have an empirical one-tailed $p \approx 0.025$ (5 of
200 permutations produce at least as many features). The result is consistent
with the original 50-permutation estimate (mean 0.00, maximum 0), and the
per-path ranking tests use 1{,}000 iterations (Methods) for higher precision.
Read with the power analysis, the pipeline is stringent (few null features
pass) but sensitive only to large effects at this sample size.

\subsection*{Ground-truth simulation confirms consensus precision}
To evaluate whether the consensus improves over single pipelines, we
constructed a synthetic metabolomics dataset (3{,}000 features, 3 groups,
$n=10$ per group) with 30 known differential features (fold changes 5--20,
analytical CV 15\%). The multiverse pipeline was applied to this ground-truth
data; the precision, recall and F1 for each
single path and the consensus are summarised in Figure~\ref{fig:simulation}.

Individual paths achieved recall of 30--40\% and F1 of 0.43--0.56. The
consensus ($\geq 2/4$ paths) achieved recall of 33\% and F1 of 0.47, with
precision of 0.77. The consensus did not dominate every single path on F1,
consistent with the conservative design that trades sensitivity for robustness.
Crucially, the consensus produced zero false positives that no single path
identified, meaning it did not introduce spurious findings. The primary value
of the multiverse remains transparency: the audit ledger shows which features
are robust across which preprocessing choices, rather than claiming that the
consensus shortlist is universally superior.

\subsection*{A signal figure confirms the effect}
The raw intensity of the top-ranked consensus
feature, Ganglioside GD1a/GD1b, separates the five cell lines clearly
(Figure~\ref{fig:ganglioside}), consistent with the extreme effect size
($\eta^2 \geq 0.94$ across all four paths).

\subsection*{Biological coherence, with two instructive failures}
Of the 15 consensus features, 13 are endogenous candidates and 12 of those are
unidentified at MSI Level~4 (no database match within ppm/score criteria), a
database-coverage limitation rather than a pipeline error. The 12 MSI~L4
features span a mass range of 583--1{,}761~Da and retention times of
4.0--10.8~min; 10 of 12 have $m/z > 600$ Da, consistent with large lipids
(glycolipids, phospholipids) that are underrepresented in standard MS/MS
libraries but are biologically plausible in a breast-cancer cell-line model.
All 12 are flagged by Path~C (MNAR-aware) and 9 of 12 by Path~A (Classical),
while only 3 are flagged by Path~B (Robust-ML) and 1 by Path~D (Middle),
reflecting the path-correlation structure discussed above. The one endogenous, identified,
all-paths feature is annotated as Ganglioside GD1a/GD1b (neutral mass 1918.06~Da,
RT 10.76~min; MSI Level~3). Gangliosides are biologically plausible discriminators
across breast-cancer subtypes, and 13 of 15 features are large-mass lipids. We
present this as evidence that the method surfaces biologically plausible signal,
\emph{not} as a validated biomarker (see Limitations). Two consensus features are
xenobiotic contaminants that the ontology-based origin filter (Stage~10) did not
catch: \texttt{2.37\_343.0578n} (a halogenated synthetic compound) and
\texttt{2.37\_703.0737m/z} (matching the NSAID Meloxicam as a 2M+H homodimer),
both identified by manual expert review. We keep them in
Table~\ref{tab:shortlist}, flagged and excluded from the endogenous count: they
show the honest failure mode of an ontology-based filter and motivate a structural
rule as future work. This is the value of an auditable pipeline: the contaminants
are visible and traceable rather than silently reported as biology.

\section*{Discussion}
The central result (Figure~\ref{fig:instability}) is that two equally defensible
preprocessing pipelines can agree on as little as one feature in twenty. Any
single-pipeline metabolomics shortlist therefore carries an unreported contingency
on choices the analyst did not vary. Running the analysis as a multiverse and
reporting only recurrent features converts that contingency into an explicit,
quantified output; combined with the per-feature audit trail, it makes the entire
path from 30{,}370 raw features to a 15-feature shortlist inspectable.

The individual components are standard: QC-RSC, SERRF, PQN, MSTUS, glog and Pareto
scaling are established methods, and platforms such as MetaboAnalyst and XCMS
implement many of them. What we add is the \emph{systematic variation of the
preprocessing pipeline as a factor}, with consensus reporting and a complete audit
ledger, rather than any new normalisation or test. This applies the multiverse and
specification-curve idea \cite{steegen2016,simonsohn2020,patel2015} to untargeted
metabolomics feature selection. A head-to-head benchmark against a fixed
single-pipeline tool would strengthen the case and is a natural next step.
The ground-truth simulation (Figure~\ref{fig:simulation}) confirms that the
consensus does not universally outperform every single path on F1, but it
adds transparency that no single path can provide. 

\textbf{Limitations.}
\emph{Single batch}: all injections were one instrument day; within-run drift is
corrected but no between-batch replicates exist.
\emph{Replication structure}: a diagnostic comparing within-group to analytical
variance was ambiguous (median ratio 1.51); whether the 6--7 replicates per line
are independent biological cultures can only be confirmed by the wet-lab protocol,
and biomarker claims require a biologically-replicated follow-up.
\emph{Power}: at $n=6$--7 only $\eta^2 > 0.5$ effects are well powered; moderate
effects are likely missed.
\emph{Identification}: 12 of 15 consensus features are MSI L4, pending targeted
MS/MS.
\emph{Origin filter gaps}: two xenobiotics passed Stage~10; a structural rule is
needed.
\emph{Path non-independence}: leave-one-path-out shows the consensus leans on
Paths~A and~C; the path set is a design choice, not an exhaustive sampling.
\emph{Deferred sensitivity}: the pipeline imputes before drift correction (reverse
of the more common order). Because QC samples have no missing values, the drift
model is unaffected by imputation in principle. However, missing-value patterns
in biological samples can alter the covariance structure of the feature matrix,
and drift correction applied after imputation may propagate imputation artefacts.
We therefore treat the current ordering as a default that warrants formal
sensitivity testing; a re-run with imputation after drift correction is listed
as future work.

\textbf{Future work.} (1) A quantitative benchmark against established tools
such as MetaboAnalyst \cite{pang2021}, IP4M \cite{tautenhahn2012} and
MetaboMSLCC \cite{boccard2022} on public datasets. (2) An imputation-order
sensitivity re-run. (3) A structural xenobiotic rule for Stage~10. (4) MS/MS
confirmation and independent-cohort validation, powered for
$\eta^2 \approx 0.14$ ($n \geq 15$ per group). (5) Exploration of weighted
voting or Bayesian model averaging as alternatives to simple-majority
consensus.

\section*{Conclusion}
Feature selection in untargeted LC--MS metabolomics is strongly dependent on
preprocessing choices that are rarely varied or reported. We described a pipeline
that treats those choices as an experimental factor, reports only
preprocessing-robust features through a tiered consensus, and records every
filtering decision in an auditable ledger. On real data, single-pipeline
shortlists overlapped by as little as Jaccard $=0.05$, whereas the multiverse
consensus was small (15 features), stringent (no false discoveries in 50 permutations) and traceable
end to end. The consensus depends on which paths are included (two correlated
paths dominate). We report the tier structure and per-path flags so that
readers can apply their own path weighting. The approach trades sensitivity for
reproducibility and transparency, an appropriate trade for the discovery phase.

\begin{backmatter}

\section*{Availability of data and materials}
Raw Progenesis QI CSV exports and Waters \_HEADER.TXT logs are available from the
corresponding author on reasonable request; a MetaboLights deposit is planned on
acceptance. Derived Parquet tables, audit ledgers and HTML reports are available
at \url{https://github.com/Ahmet-Kaplan/metabolomic\_pipeline} under
\texttt{data/}. Supporting analyses and figures regenerate
via \texttt{scripts/diagnose\_multiverse\_instability.py},
\texttt{scripts/build\_paper\_tables.py} and
\texttt{scripts/make\_paper\_figures.py}.

\section*{Availability of code}
Open source under the MIT License at
\url{https://github.com/Ahmet-Kaplan/metabolomic\_pipeline}; the version
producing these results is archived on Zenodo
(\href{https://doi.org/10.5281/zenodo.21443824}{10.5281/zenodo.21443824}).

\section*{Competing interests}
The authors declare no competing interests.

\section*{Authors' contributions}
IY and MSA designed and implemented the pipeline. AK performed the analyses and wrote the
manuscript. All authors read and approved the final manuscript.

\section*{Acknowledgements}
The authors thank their research advisor and the domain expert for guidance on the
filter cascade design and threshold selection.

\section*{Ethics}
No ethics approval was required; all cell lines were obtained from commercial
repositories (ATCC) and are not human-subjects research.

\bibliographystyle{bmc-mathphys}

\newcommand{\BMCxmlcomment}[1]{}

\BMCxmlcomment{

<refgrp>

<bibl id="B1">
  <title><p>Guidelines and considerations for the use of system suitability and
  quality control samples in mass spectrometry assays applied in untargeted
  clinical metabolomic studies</p></title>
  <aug>
    <au><snm>Broadhurst</snm><fnm>D</fnm></au>
    <au><snm>Goodacre</snm><fnm>R</fnm></au>
    <au><snm>Reinke</snm><fnm>SN</fnm></au>
    <au><snm>Kuligowski</snm><fnm>J</fnm></au>
    <au><snm>Wilson</snm><fnm>ID</fnm></au>
    <au><snm>Lewis</snm><fnm>MR</fnm></au>
    <au><snm>Dunn</snm><fnm>WB</fnm></au>
  </aug>
  <source>Metabolomics</source>
  <pubdate>2018</pubdate>
  <volume>14</volume>
  <fpage>72</fpage>
</bibl>

<bibl id="B2">
  <title><p>Statistical strategies for avoiding false discoveries in
  metabolomics and related experiments</p></title>
  <aug>
    <au><snm>Broadhurst</snm><fnm>DI</fnm></au>
    <au><snm>Kell</snm><fnm>DB</fnm></au>
  </aug>
  <source>Metabolomics</source>
  <pubdate>2006</pubdate>
  <volume>2</volume>
  <issue>4</issue>
  <fpage>171</fpage>
  <lpage>-196</lpage>
</bibl>

<bibl id="B3">
  <title><p>Increasing Transparency Through a Multiverse Analysis</p></title>
  <aug>
    <au><snm>Steegen</snm><fnm>S</fnm></au>
    <au><snm>Tuerlinckx</snm><fnm>F</fnm></au>
    <au><snm>Gelman</snm><fnm>A</fnm></au>
    <au><snm>Vanpaemel</snm><fnm>W</fnm></au>
  </aug>
  <source>Perspectives on Psychological Science</source>
  <pubdate>2016</pubdate>
  <volume>11</volume>
  <issue>5</issue>
  <fpage>702</fpage>
  <lpage>-712</lpage>
</bibl>

<bibl id="B4">
  <title><p>Specification curve analysis</p></title>
  <aug>
    <au><snm>Simonsohn</snm><fnm>U</fnm></au>
    <au><snm>Simmons</snm><fnm>JP</fnm></au>
    <au><snm>Nelson</snm><fnm>LD</fnm></au>
  </aug>
  <source>Nature Human Behaviour</source>
  <pubdate>2020</pubdate>
  <volume>4</volume>
  <fpage>1208</fpage>
  <lpage>-1214</lpage>
</bibl>

<bibl id="B5">
  <title><p>Assessment of vibration of effects due to model specification can
  demonstrate the instability of observational associations</p></title>
  <aug>
    <au><snm>Patel</snm><fnm>CJ</fnm></au>
    <au><snm>Burford</snm><fnm>B</fnm></au>
    <au><snm>Ioannidis</snm><fnm>JPA</fnm></au>
  </aug>
  <source>Journal of Clinical Epidemiology</source>
  <pubdate>2015</pubdate>
  <volume>68</volume>
  <issue>9</issue>
  <fpage>1046</fpage>
  <lpage>-1058</lpage>
</bibl>

<bibl id="B6">
  <title><p>Leakage and the reproducibility crisis in machine-learning-based
  science</p></title>
  <aug>
    <au><snm>Kapoor</snm><fnm>S</fnm></au>
    <au><snm>Narayanan</snm><fnm>A</fnm></au>
  </aug>
  <source>Patterns</source>
  <pubdate>2023</pubdate>
  <volume>4</volume>
  <issue>9</issue>
  <fpage>100804</fpage>
</bibl>

<bibl id="B7">
  <title><p>Probabilistic quotient normalization as robust method to account
  for dilution of complex biological mixtures</p></title>
  <aug>
    <au><snm>Dieterle</snm><fnm>F</fnm></au>
    <au><snm>Ross</snm><fnm>A</fnm></au>
    <au><snm>Schlotterbeck</snm><fnm>G</fnm></au>
    <au><snm>Senn</snm><fnm>H</fnm></au>
  </aug>
  <source>Analytical Chemistry</source>
  <pubdate>2006</pubdate>
  <volume>78</volume>
  <issue>13</issue>
  <fpage>4281</fpage>
  <lpage>-4290</lpage>
</bibl>

<bibl id="B8">
  <title><p>MissForest---non-parametric missing value imputation for mixed-type
  data</p></title>
  <aug>
    <au><snm>Stekhoven</snm><fnm>DJ</fnm></au>
    <au><snm>B{\"u}hlmann</snm><fnm>P</fnm></au>
  </aug>
  <source>Bioinformatics</source>
  <pubdate>2012</pubdate>
  <volume>28</volume>
  <issue>1</issue>
  <fpage>112</fpage>
  <lpage>-118</lpage>
</bibl>

<bibl id="B9">
  <title><p>Accounting for the multiple natures of missing values in label-free
  quantitative proteomics data sets to compare imputation
  strategies</p></title>
  <aug>
    <au><snm>Lazar</snm><fnm>C</fnm></au>
    <au><snm>Gatto</snm><fnm>L</fnm></au>
    <au><snm>Ferro</snm><fnm>M</fnm></au>
    <au><snm>Bruley</snm><fnm>C</fnm></au>
    <au><snm>Burger</snm><fnm>T</fnm></au>
  </aug>
  <source>Journal of Proteome Research</source>
  <pubdate>2016</pubdate>
  <volume>15</volume>
  <issue>4</issue>
  <fpage>1116</fpage>
  <lpage>-1125</lpage>
</bibl>

<bibl id="B10">
  <title><p>Procedures for large-scale metabolic profiling of serum and plasma
  using gas chromatography and liquid chromatography coupled to mass
  spectrometry</p></title>
  <aug>
    <au><snm>Dunn</snm><fnm>WB</fnm></au>
    <au><cnm>others</cnm></au>
  </aug>
  <source>Nature Protocols</source>
  <pubdate>2011</pubdate>
  <volume>6</volume>
  <issue>7</issue>
  <fpage>1060</fpage>
  <lpage>-1083</lpage>
</bibl>

<bibl id="B11">
  <title><p>Systematic error removal using random forest for normalizing
  large-scale untargeted lipidomics data</p></title>
  <aug>
    <au><snm>Fan</snm><fnm>S</fnm></au>
    <au><snm>Kind</snm><fnm>T</fnm></au>
    <au><snm>Cajka</snm><fnm>T</fnm></au>
    <au><snm>Hazen</snm><fnm>SL</fnm></au>
    <au><snm>Tang</snm><fnm>WHW</fnm></au>
    <au><snm>Kaddurah Daouk</snm><fnm>R</fnm></au>
    <au><snm>Irvin</snm><fnm>MR</fnm></au>
    <au><snm>Arnett</snm><fnm>DK</fnm></au>
    <au><snm>Barupal</snm><fnm>DK</fnm></au>
    <au><snm>Fiehn</snm><fnm>O</fnm></au>
  </aug>
  <source>Analytical Chemistry</source>
  <pubdate>2019</pubdate>
  <volume>91</volume>
  <issue>5</issue>
  <fpage>3590</fpage>
  <lpage>-3596</lpage>
</bibl>

<bibl id="B12">
  <title><p>A note on normalization of biofluid 1D 1H-NMR data</p></title>
  <aug>
    <au><snm>Torgrip</snm><fnm>R J O</fnm></au>
    <au><snm>{\AA}berg</snm><fnm>K M</fnm></au>
    <au><snm>Alm</snm><fnm>E</fnm></au>
    <au><snm>Schuppe Koistinen</snm><fnm>I</fnm></au>
    <au><snm>Lindberg</snm><fnm>J</fnm></au>
  </aug>
  <source>Metabolomics</source>
  <pubdate>2008</pubdate>
  <volume>4</volume>
  <fpage>114</fpage>
  <lpage>-121</lpage>
</bibl>

<bibl id="B13">
  <title><p>MetaboAnalyst 5.0: narrowing the gap between raw spectra and
  functional insights</p></title>
  <aug>
    <au><snm>Pang</snm><fnm>Z</fnm></au>
    <au><snm>Chong</snm><fnm>J</fnm></au>
    <au><snm>Zhou</snm><fnm>G</fnm></au>
    <au><snm>Lima Morais</snm><fnm>DA</fnm></au>
    <au><snm>Chang</snm><fnm>L</fnm></au>
    <au><snm>Barrette</snm><fnm>M</fnm></au>
    <au><snm>Gauthier</snm><fnm>C</fnm></au>
    <au><snm>Jacques</snm><fnm>P{\'E}</fnm></au>
    <au><snm>Li</snm><fnm>S</fnm></au>
    <au><snm>Xia</snm><fnm>J</fnm></au>
  </aug>
  <source>Nucleic Acids Research</source>
  <pubdate>2021</pubdate>
  <volume>49</volume>
  <issue>W1</issue>
  <fpage>W388</fpage>
  <lpage>-W396</lpage>
</bibl>

<bibl id="B14">
  <title><p>XCMS: processing mass spectrometry data for metabolite profiling
  using nonlinear peak alignment, matching, and identification</p></title>
  <aug>
    <au><snm>Smith</snm><fnm>CA</fnm></au>
    <au><snm>Want</snm><fnm>EJ</fnm></au>
    <au><snm>O'Maille</snm><fnm>G</fnm></au>
    <au><snm>Abagyan</snm><fnm>R</fnm></au>
    <au><snm>Siuzdak</snm><fnm>G</fnm></au>
  </aug>
  <source>Analytical Chemistry</source>
  <pubdate>2006</pubdate>
  <volume>78</volume>
  <issue>3</issue>
  <fpage>779</fpage>
  <lpage>-787</lpage>
</bibl>

<bibl id="B15">
  <title><p>notame: workflow for non-targeted LC--MS metabolic
  profiling</p></title>
  <aug>
    <au><snm>Klavus</snm><fnm>A</fnm></au>
    <au><snm>Kokla</snm><fnm>M</fnm></au>
    <au><snm>Noerman</snm><fnm>S</fnm></au>
    <au><snm>Koistinen</snm><fnm>VM</fnm></au>
    <au><snm>Tuomainen</snm><fnm>M</fnm></au>
    <au><snm>Zarei</snm><fnm>I</fnm></au>
    <au><snm>Meuronen</snm><fnm>T</fnm></au>
    <au><snm>H{\"a}kkinen</snm><fnm>MR</fnm></au>
    <au><snm>Rummukainen</snm><fnm>S</fnm></au>
    <au><snm>Farizah Babu</snm><fnm>A</fnm></au>
    <au><snm>Sallinen</snm><fnm>T</fnm></au>
    <au><snm>K{\"a}rr{\"a}nen</snm><fnm>M</fnm></au>
    <au><snm>Hanhineva</snm><fnm>K</fnm></au>
    <au><snm>Brunius</snm><fnm>C</fnm></au>
  </aug>
  <source>Metabolites</source>
  <pubdate>2020</pubdate>
  <volume>10</volume>
  <issue>4</issue>
  <fpage>158</fpage>
</bibl>

<bibl id="B16">
  <title><p>IPO: a tool for automated optimization of XCMS
  parameters</p></title>
  <aug>
    <au><snm>Libiseller</snm><fnm>G</fnm></au>
    <au><snm>Dvorzak</snm><fnm>M</fnm></au>
    <au><snm>Kleb</snm><fnm>U</fnm></au>
    <au><snm>Gander</snm><fnm>D</fnm></au>
    <au><snm>Eisenberg</snm><fnm>T</fnm></au>
    <au><snm>Madeo</snm><fnm>F</fnm></au>
    <au><snm>Neumann</snm><fnm>S</fnm></au>
    <au><snm>Trausinger</snm><fnm>G</fnm></au>
    <au><snm>Sinner</snm><fnm>F</fnm></au>
    <au><snm>Pieber</snm><fnm>TR</fnm></au>
    <au><snm>Magnes</snm><fnm>C</fnm></au>
  </aug>
  <source>BMC Bioinformatics</source>
  <pubdate>2015</pubdate>
  <volume>16</volume>
  <fpage>118</fpage>
</bibl>

<bibl id="B17">
  <title><p>Stability selection</p></title>
  <aug>
    <au><snm>Meinshausen</snm><fnm>N</fnm></au>
    <au><snm>B{\"u}hlmann</snm><fnm>P</fnm></au>
  </aug>
  <source>Journal of the Royal Statistical Society: Series B (Statistical
  Methodology)</source>
  <pubdate>2010</pubdate>
  <volume>72</volume>
  <issue>4</issue>
  <fpage>417</fpage>
  <lpage>-473</lpage>
</bibl>

<bibl id="B18">
  <title><p>Robust feature selection using ensemble feature selection
  techniques</p></title>
  <aug>
    <au><snm>Saeys</snm><fnm>Y</fnm></au>
    <au><snm>Abeel</snm><fnm>T</fnm></au>
    <au><snm>Peer</snm><fnm>Y</fnm></au>
  </aug>
  <source>Machine Learning and Knowledge Discovery in Databases (ECML
  PKDD)</source>
  <pubdate>2008</pubdate>
  <volume>5212</volume>
  <fpage>313</fpage>
  <lpage>-325</lpage>
</bibl>

<bibl id="B19">
  <title><p>Improving Reproducibility in Machine Learning Research</p></title>
  <aug>
    <au><snm>Pineau</snm><fnm>J</fnm></au>
    <au><snm>Vincent Lamarre</snm><fnm>P</fnm></au>
    <au><snm>Sinha</snm><fnm>K</fnm></au>
    <au><snm>Larivi{\`e}re</snm><fnm>V</fnm></au>
    <au><snm>Beygelzimer</snm><fnm>A</fnm></au>
    <au><snm>Buc</snm><fnm>F</fnm></au>
    <au><snm>Fox</snm><fnm>E</fnm></au>
    <au><snm>Larochelle</snm><fnm>H</fnm></au>
  </aug>
  <source>Journal of Machine Learning Research</source>
  <pubdate>2021</pubdate>
  <volume>22</volume>
  <issue>164</issue>
  <fpage>1</fpage>
  <lpage>-20</lpage>
  <url>https://jmlr.org/papers/v22/20-303.html</url>
</bibl>

<bibl id="B20">
  <title><p>Navigating the pitfalls of applying machine learning in
  genomics</p></title>
  <aug>
    <au><snm>Whalen</snm><fnm>S</fnm></au>
    <au><snm>Schreiber</snm><fnm>J</fnm></au>
    <au><snm>Noble</snm><fnm>WS</fnm></au>
    <au><snm>Pollard</snm><fnm>KS</fnm></au>
  </aug>
  <source>Nature Reviews Genetics</source>
  <pubdate>2022</pubdate>
  <volume>23</volume>
  <fpage>169</fpage>
  <lpage>-181</lpage>
</bibl>

<bibl id="B21">
  <title><p>Reproducibility standards for machine learning in the life
  sciences</p></title>
  <aug>
    <au><snm>Heil</snm><fnm>BJ</fnm></au>
    <au><snm>Hoffman</snm><fnm>MM</fnm></au>
    <au><snm>Markowetz</snm><fnm>F</fnm></au>
    <au><snm>Lee</snm><fnm>SI</fnm></au>
    <au><snm>Greene</snm><fnm>CS</fnm></au>
    <au><snm>Hicks</snm><fnm>SC</fnm></au>
  </aug>
  <source>Nature Methods</source>
  <pubdate>2021</pubdate>
  <volume>18</volume>
  <fpage>1132</fpage>
  <lpage>-1135</lpage>
</bibl>

<bibl id="B22">
  <title><p>Assessment of PLSDA cross validation</p></title>
  <aug>
    <au><snm>Westerhuis</snm><fnm>JA</fnm></au>
    <au><snm>Hoefsloot</snm><fnm>HCJ</fnm></au>
    <au><snm>Smit</snm><fnm>S</fnm></au>
    <au><snm>Vis</snm><fnm>DJ</fnm></au>
    <au><snm>Smilde</snm><fnm>AK</fnm></au>
    <au><snm>Velzen</snm><fnm>EJJ</fnm></au>
    <au><snm>Duijnhoven</snm><fnm>JPM</fnm></au>
    <au><snm>Dorsten</snm><fnm>FA</fnm></au>
  </aug>
  <source>Metabolomics</source>
  <pubdate>2008</pubdate>
  <volume>4</volume>
  <issue>1</issue>
  <fpage>81</fpage>
  <lpage>-89</lpage>
</bibl>

<bibl id="B23">
  <title><p>Sparse PLS discriminant analysis: biologically relevant feature
  selection and graphical displays for multiclass problems</p></title>
  <aug>
    <au><snm>L{\^e} Cao</snm><fnm>KA</fnm></au>
    <au><snm>Boitard</snm><fnm>S</fnm></au>
    <au><snm>Besse</snm><fnm>P</fnm></au>
  </aug>
  <source>BMC Bioinformatics</source>
  <pubdate>2011</pubdate>
  <volume>12</volume>
  <fpage>253</fpage>
</bibl>

<bibl id="B24">
  <title><p>Random forests</p></title>
  <aug>
    <au><snm>Breiman</snm><fnm>L</fnm></au>
  </aug>
  <source>Machine Learning</source>
  <pubdate>2001</pubdate>
  <volume>45</volume>
  <fpage>5</fpage>
  <lpage>-32</lpage>
</bibl>

<bibl id="B25">
  <title><p>Missing value estimation methods for DNA microarrays</p></title>
  <aug>
    <au><snm>Troyanskaya</snm><fnm>O</fnm></au>
    <au><snm>Cantor</snm><fnm>M</fnm></au>
    <au><snm>Sherlock</snm><fnm>G</fnm></au>
    <au><snm>Brown</snm><fnm>P</fnm></au>
    <au><snm>Hastie</snm><fnm>T</fnm></au>
    <au><snm>Tibshirani</snm><fnm>R</fnm></au>
    <au><snm>Botstein</snm><fnm>D</fnm></au>
    <au><snm>Altman</snm><fnm>RB</fnm></au>
  </aug>
  <source>Bioinformatics</source>
  <pubdate>2001</pubdate>
  <volume>17</volume>
  <issue>6</issue>
  <fpage>520</fpage>
  <lpage>-525</lpage>
</bibl>

<bibl id="B26">
  <title><p>Improved classification accuracy in 1- and 2-dimensional NMR
  metabolomics data using the variance stabilising generalised logarithm
  transformation</p></title>
  <aug>
    <au><snm>Parsons</snm><fnm>HM</fnm></au>
    <au><snm>Ludwig</snm><fnm>C</fnm></au>
    <au><snm>G{\"u}nther</snm><fnm>UL</fnm></au>
    <au><snm>Viant</snm><fnm>MR</fnm></au>
  </aug>
  <source>BMC Bioinformatics</source>
  <pubdate>2007</pubdate>
  <volume>8</volume>
  <fpage>234</fpage>
</bibl>

<bibl id="B27">
  <title><p>Centering, scaling, and transformations: improving the biological
  information content of metabolomics data</p></title>
  <aug>
    <au><snm>Berg</snm><fnm>RA</fnm></au>
    <au><snm>Hoefsloot</snm><fnm>HCJ</fnm></au>
    <au><snm>Westerhuis</snm><fnm>JA</fnm></au>
    <au><snm>Smilde</snm><fnm>AK</fnm></au>
    <au><snm>Werf</snm><fnm>MJ</fnm></au>
  </aug>
  <source>BMC Genomics</source>
  <pubdate>2006</pubdate>
  <volume>7</volume>
  <fpage>142</fpage>
</bibl>

<bibl id="B28">
  <title><p>An accelerated workflow for untargeted metabolomics using the
  METLIN database</p></title>
  <aug>
    <au><snm>Tautenhahn</snm><fnm>R</fnm></au>
    <au><snm>Cho</snm><fnm>K</fnm></au>
    <au><snm>Uritboonthai</snm><fnm>W</fnm></au>
    <au><snm>Zhu</snm><fnm>Z</fnm></au>
    <au><snm>Patti</snm><fnm>GJ</fnm></au>
    <au><snm>Siuzdak</snm><fnm>G</fnm></au>
  </aug>
  <source>Nature Biotechnology</source>
  <pubdate>2012</pubdate>
  <volume>30</volume>
  <fpage>826</fpage>
  <lpage>-828</lpage>
</bibl>

<bibl id="B29">
  <title><p>MetaboMSLCC: a multi-block statistical approach for the analysis of
  metabolomics data</p></title>
  <aug>
    <au><snm>Boccard</snm><fnm>J</fnm></au>
    <au><snm>Badoud</snm><fnm>F</fnm></au>
    <au><snm>Grata</snm><fnm>E</fnm></au>
    <au><snm>Ouertani</snm><fnm>S</fnm></au>
    <au><snm>Hanafi</snm><fnm>M</fnm></au>
    <au><snm>Mazerolles</snm><fnm>G</fnm></au>
    <au><snm>Lalibert{\'e}</snm><fnm>A</fnm></au>
    <au><snm>Veuthey</snm><fnm>JL</fnm></au>
    <au><snm>Lanfumey</snm><fnm>L</fnm></au>
    <au><snm>Rudaz</snm><fnm>S</fnm></au>
  </aug>
  <source>Metabolomics</source>
  <pubdate>2022</pubdate>
  <volume>18</volume>
  <fpage>45</fpage>
</bibl>

<bibl id="B30">
  <title><p>Large-scale human metabolomics studies: a strategy for data
  (pre-)processing and validation</p></title>
  <aug>
    <au><snm>Bijlsma</snm><fnm>S</fnm></au>
    <au><snm>Bobeldijk</snm><fnm>I</fnm></au>
    <au><snm>Verheij</snm><fnm>ER</fnm></au>
    <au><snm>Ramaker</snm><fnm>R</fnm></au>
    <au><snm>Kochhar</snm><fnm>S</fnm></au>
    <au><snm>Macdonald</snm><fnm>IA</fnm></au>
    <au><snm>Ommen</snm><fnm>B</fnm></au>
    <au><snm>Smilde</snm><fnm>AK</fnm></au>
  </aug>
  <source>Analytical Chemistry</source>
  <pubdate>2006</pubdate>
  <volume>78</volume>
  <issue>2</issue>
  <fpage>567</fpage>
  <lpage>-574</lpage>
</bibl>

<bibl id="B31">
  <title><p>Development and application of ultra-performance liquid
  chromatography-TOF MS for precision large scale urinary metabolic
  phenotyping</p></title>
  <aug>
    <au><snm>Lewis</snm><fnm>MR</fnm></au>
    <au><cnm>others</cnm></au>
  </aug>
  <source>Analytical Chemistry</source>
  <pubdate>2016</pubdate>
  <volume>88</volume>
  <issue>18</issue>
  <fpage>9004</fpage>
  <lpage>-9013</lpage>
</bibl>

<bibl id="B32">
  <title><p>Global metabolic profiling procedures for urine using
  UPLC-MS</p></title>
  <aug>
    <au><snm>Want</snm><fnm>EJ</fnm></au>
    <au><snm>Wilson</snm><fnm>ID</fnm></au>
    <au><snm>Gika</snm><fnm>H</fnm></au>
    <au><snm>Theodoridis</snm><fnm>G</fnm></au>
    <au><snm>Plumb</snm><fnm>RS</fnm></au>
    <au><snm>Shockcor</snm><fnm>J</fnm></au>
    <au><snm>Holmes</snm><fnm>E</fnm></au>
    <au><snm>Nicholson</snm><fnm>JK</fnm></au>
  </aug>
  <source>Nature Protocols</source>
  <pubdate>2010</pubdate>
  <volume>5</volume>
  <issue>6</issue>
  <fpage>1005</fpage>
  <lpage>-1018</lpage>
</bibl>

<bibl id="B33">
  <title><p>Collision cross section compendium to annotate and predict
  multi-omic compound identities</p></title>
  <aug>
    <au><snm>Picache</snm><fnm>JA</fnm></au>
    <au><snm>Rose</snm><fnm>BS</fnm></au>
    <au><snm>Balinski</snm><fnm>A</fnm></au>
    <au><snm>Leaptrot</snm><fnm>KL</fnm></au>
    <au><snm>Sherrod</snm><fnm>SD</fnm></au>
    <au><snm>May</snm><fnm>JC</fnm></au>
    <au><snm>McLean</snm><fnm>JA</fnm></au>
  </aug>
  <source>Chemical Science</source>
  <pubdate>2019</pubdate>
  <volume>10</volume>
  <fpage>983</fpage>
  <lpage>-993</lpage>
</bibl>

<bibl id="B34">
  <title><p>Proposed minimum reporting standards for chemical
  analysis</p></title>
  <aug>
    <au><snm>Sumner</snm><fnm>LW</fnm></au>
    <au><cnm>others</cnm></au>
  </aug>
  <source>Metabolomics</source>
  <pubdate>2007</pubdate>
  <volume>3</volume>
  <issue>3</issue>
  <fpage>211</fpage>
  <lpage>-221</lpage>
</bibl>

<bibl id="B35">
  <title><p>Identifying small molecules via high resolution mass spectrometry:
  communicating confidence</p></title>
  <aug>
    <au><snm>Schymanski</snm><fnm>EL</fnm></au>
    <au><snm>Jeon</snm><fnm>J</fnm></au>
    <au><snm>Gulde</snm><fnm>R</fnm></au>
    <au><snm>Fenner</snm><fnm>K</fnm></au>
    <au><snm>Ruff</snm><fnm>M</fnm></au>
    <au><snm>Singer</snm><fnm>HP</fnm></au>
    <au><snm>Hollender</snm><fnm>J</fnm></au>
  </aug>
  <source>Environmental Science \& Technology</source>
  <pubdate>2014</pubdate>
  <volume>48</volume>
  <issue>4</issue>
  <fpage>2097</fpage>
  <lpage>-2098</lpage>
</bibl>

<bibl id="B36">
  <title><p>HMDB 5.0: the Human Metabolome Database for 2022</p></title>
  <aug>
    <au><snm>Wishart</snm><fnm>DS</fnm></au>
    <au><cnm>others</cnm></au>
  </aug>
  <source>Nucleic Acids Research</source>
  <pubdate>2022</pubdate>
  <volume>50</volume>
  <fpage>D622</fpage>
  <lpage>-D631</lpage>
</bibl>

</refgrp>
} 

\section*{Figures}

\begin{figure}[h!]
\centering
\includegraphics[width=0.6\textwidth]{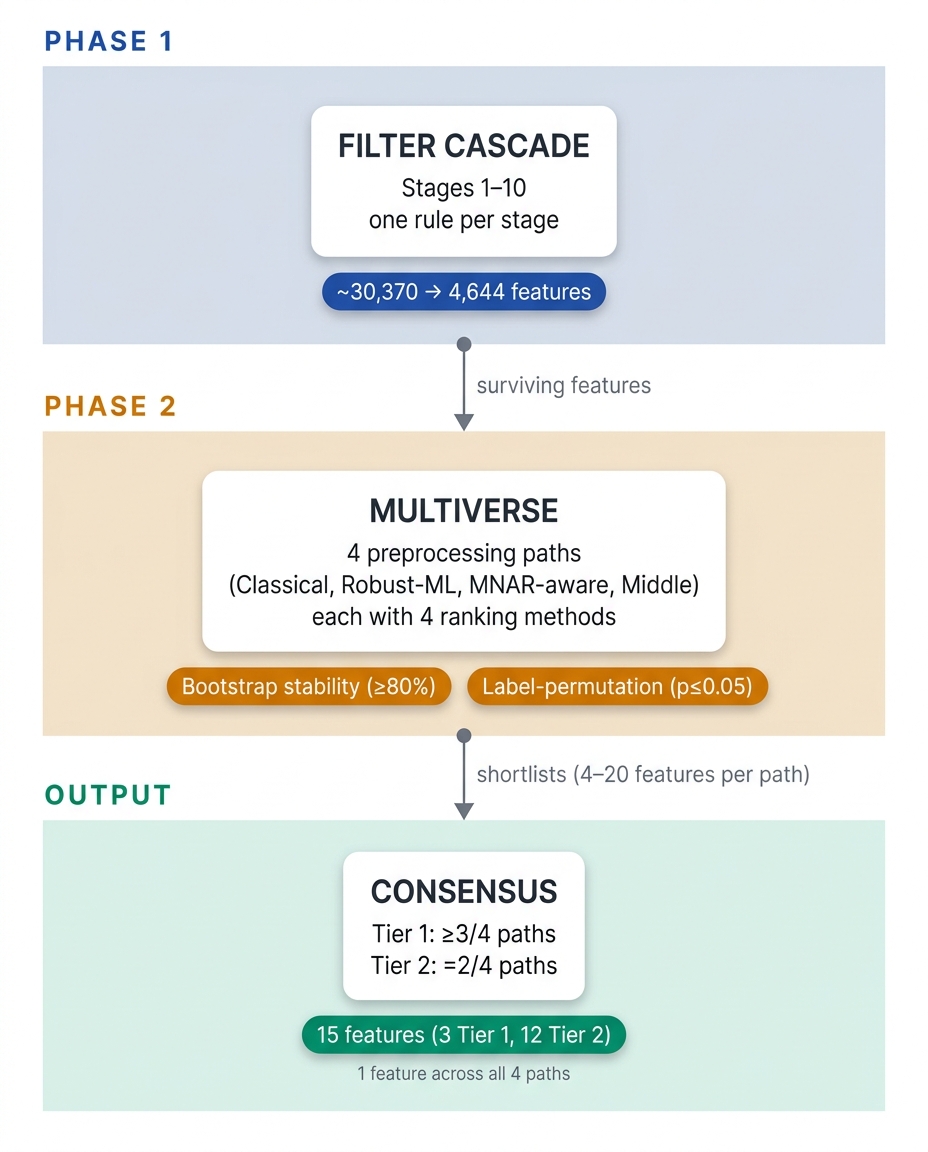}
\caption{\csentence{Pipeline schematic.}
  The auditable QC filter cascade (Phase~1) feeds the multiverse (Phase~2); only
  features recurring across preprocessing paths enter the tiered consensus.}
\label{fig:schematic}
\end{figure}

\begin{figure}[h!]
\centering
\includegraphics[width=0.8\textwidth]{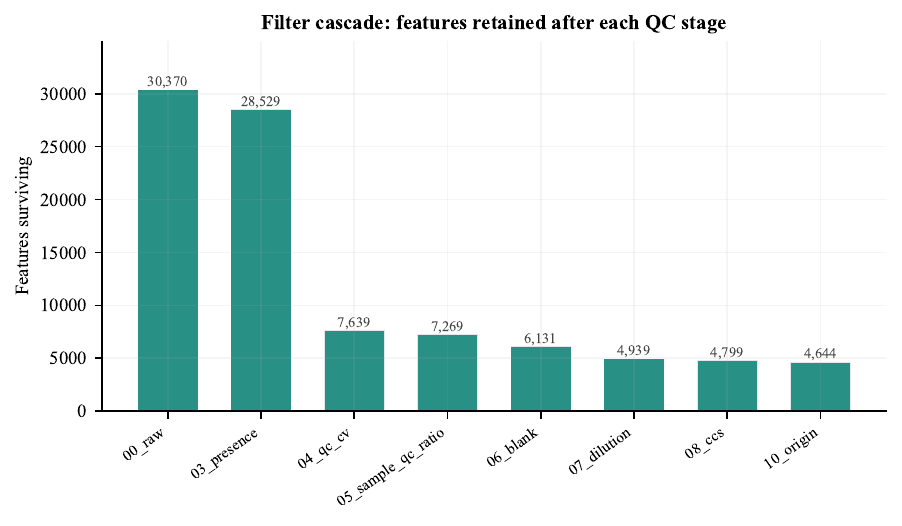}
\caption{\csentence{Filter-cascade waterfall.}
  Features surviving after each QC filter stage on the demonstration data
  (30{,}370 $\rightarrow$ 4{,}644). Stage~4 (QC-CV) is the largest single
  reduction.}
\label{fig:waterfall}
\end{figure}

\begin{figure}[h!]
\centering
\includegraphics[width=0.95\textwidth]{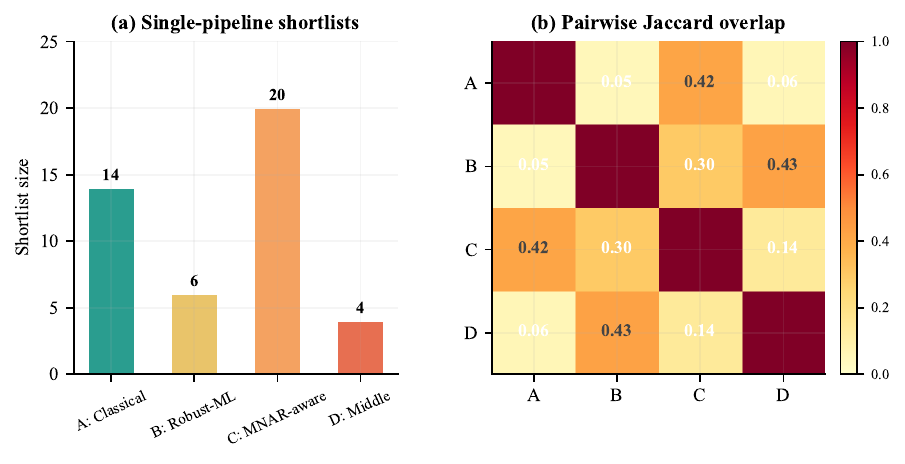}
\caption{\csentence{Single-pipeline instability.}
  (a) Shortlist size when each preprocessing path is run alone.
  (b) Pairwise Jaccard overlap between the single-pipeline shortlists (minimum
  0.05).}
\label{fig:instability}
\end{figure}

\begin{figure}[h!]
\centering
\includegraphics[width=0.75\textwidth]{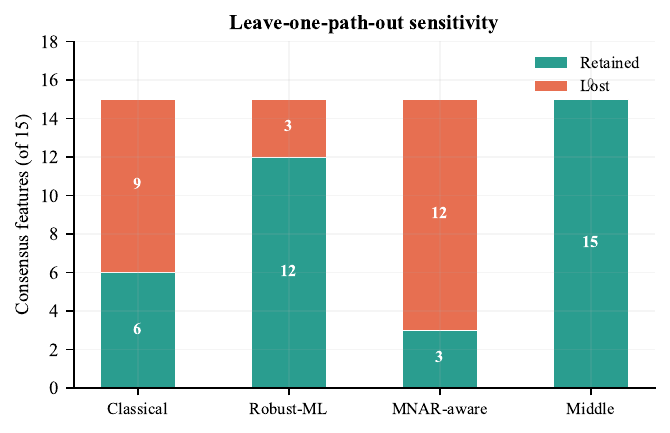}
\caption{\csentence{Leave-one-path-out robustness.}
  Consensus features (of 15) retained versus lost when each path is removed from
  the $\geq 2/4$ consensus.}
\label{fig:loo}
\end{figure}

\begin{figure}[h!]
\centering
\includegraphics[width=0.7\textwidth]{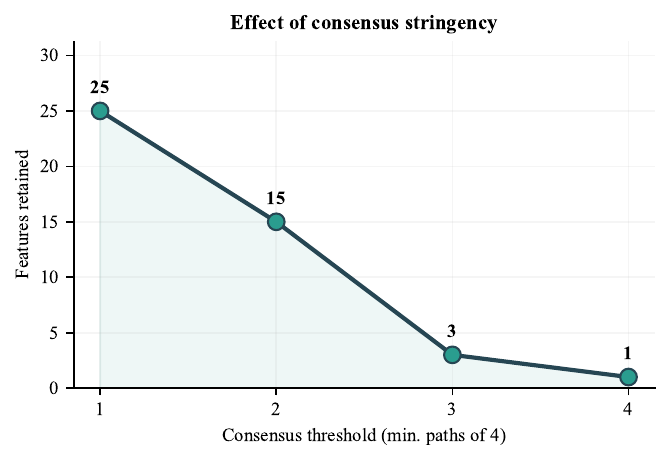}
\caption{\csentence{Consensus-threshold ablation.}
  Features retained as the consensus threshold tightens from $\geq 1$ to
  $\geq 4$ paths (25, 15, 3, 1).}
\label{fig:ablation}
\end{figure}

\begin{figure}[h!]
\centering
\includegraphics[width=0.7\textwidth]{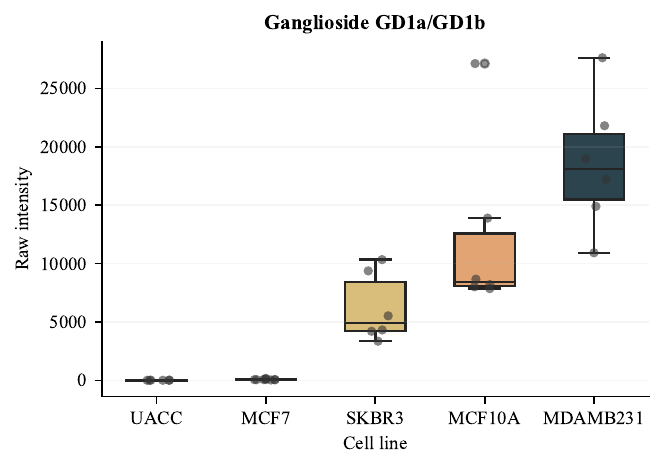}
\caption{\csentence{Ganglioside GD1a/GD1b across cell lines.}
  Raw intensity of the top-ranked consensus feature
  (\texttt{10.76\_1918.0639n}) across the five breast-cancer cell lines.
  $\eta^2\geq 0.94$ across all four preprocessing paths.}
\label{fig:ganglioside}
\end{figure}

\begin{figure}[h!]
\centering
\includegraphics[width=0.85\textwidth]{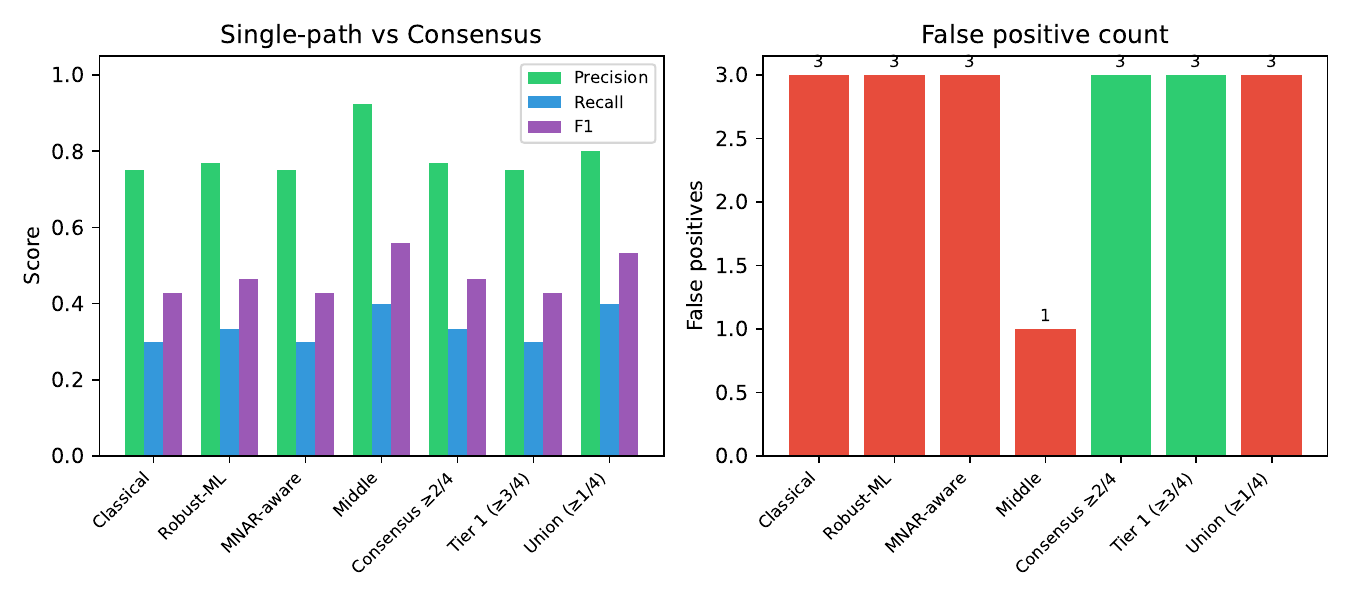}
\caption{\csentence{Ground-truth simulation benchmark.}
  Precision, recall and F1 for each single preprocessing path and the
  multiverse consensus ($\geq 2/4$ paths) on a synthetic dataset with 30
  known differential features (3{,}000 total features, 3 groups, $n=10$).
  The consensus achieves precision comparable to the best single path while
  providing the audit-trail transparency that no single path can offer.}
\label{fig:simulation}
\end{figure}

\section*{Tables}

\begin{table}[h!]
\caption{The 15-feature consensus shortlist. $q$ and $\eta^2$ are the best
(minimum $q$, maximum $\eta^2$) values reached across the four paths, an
optimistic summary; per-path values differ substantially (see machine-readable
table). MSI tier from Stage~9; xenobiotics flagged and excluded from the
endogenous count. Per-path $q$/$\eta^2$ are available in
\texttt{data/reports/multiverse/paper\_analysis/per\_path\_q\_eta.csv}.
$\eta^2_{\max}$ values of 1.00 are rounded from $>0.995$.}
\label{tab:shortlist}
\begin{tabular}{llccc}
\hline
Feature & Tier / paths & MSI & $\eta^2_{\max}$ & Note \\
\hline
10.76\_1918.0639n & Tier 1 (4/4) & L3 & 0.99 & Ganglioside GD1a/b \\
10.55\_1761.0087n & Tier 1 (3/4) & L4 & 0.97 & unidentified lipid \\
2.37\_343.0578n   & Tier 1 (3/4) & L2 & 0.99 & xenobiotic (halogenated) \\
10.78\_1702.0105n & Tier 2 (2/4) & L4 & 0.95 & unidentified \\
10.81\_1498.8843n & Tier 2 (2/4) & L4 & 0.99 & unidentified \\
2.37\_703.0737m/z & Tier 2 (2/4) & L3 & 0.99 & xenobiotic (Meloxicam) \\
4.03\_659.2403n   & Tier 2 (2/4) & L4 & 1.00 & unidentified \\
4.03\_675.2058n   & Tier 2 (2/4) & L4 & 1.00 & unidentified \\
4.03\_691.1686n   & Tier 2 (2/4) & L4 & 1.00 & unidentified \\
4.49\_583.1440n   & Tier 2 (2/4) & L4 & 1.00 & unidentified \\
4.49\_599.1149n   & Tier 2 (2/4) & L4 & 1.00 & unidentified \\
8.12\_938.3263n   & Tier 2 (2/4) & L4 & 1.00 & unidentified \\
8.13\_900.3844n   & Tier 2 (2/4) & L4 & 1.00 & unidentified \\
8.24\_1273.7281n  & Tier 2 (2/4) & L4 & 0.97 & unidentified \\
8.96\_1386.8145n  & Tier 2 (2/4) & L4 & 0.98 & unidentified \\
\hline
\end{tabular}
\end{table}

\begin{table}[h!]
\caption{Filter-cascade stages: rule, default threshold and literature source.}
\label{tab:provenance}
\begin{tabular}{llll}
\hline
Stage & Rule & Threshold & Source \\
\hline
3 Presence  & detected in $\geq 80\%$ of $\geq 1$ group & 80\% & \cite{bijlsma2006} \\
4 QC-CV     & QC-pool CV & $\leq 30\%$ & \cite{broadhurst2018} \\
5 D-ratio   & biological $>$ technical variation & ratio $>1$ & \cite{broadhurst2018} \\
6 Blank     & blank vs sample+QC mean & $\leq 20\%$ & \cite{broadhurst2018} \\
7 Dilution  & signal linear with concentration & $R^2 \geq 0.70$ & \cite{lewis2016,want2010} \\
8 CCS       & ion-mobility size vs reference & error $\leq 2$--3\% & \cite{picache2019} \\
9 ID rank   & assign MSI tier (L2/L3/L4) & annotation only & \cite{sumner2007,schymanski2014} \\
10 Origin   & drop purely-exogenous features & HMDB ontology & \cite{wishart2022} \\
\hline
\end{tabular}
\end{table}

\end{backmatter}
\end{document}